\documentclass{article} 
\usepackage[preprint]{colm2025_conference}

\usepackage{microtype}
\usepackage{hyperref}
\usepackage{url}
\usepackage{booktabs}
\usepackage{graphicx}
\usepackage{lineno}
\usepackage{wrapfig}
\usepackage[bottom]{footmisc}

\usepackage{yfonts}     
\usepackage{microtype}  
\usepackage{pifont}     
\usepackage{lineno}
\usepackage[frozencache,cachedir=minted-cache]{minted}
\usepackage{amsmath,amssymb}

\def\Snospace~{\S{}}

\definecolor{coco1}{HTML}{D9E4EC}
\definecolor{coco2}{HTML}{B7CFDC}
\definecolor{coco3}{HTML}{6AABD2}
\definecolor{coco4}{HTML}{385E72}
\hypersetup{
    colorlinks=true,
    linkcolor=coco3,
    filecolor=coco4,      
    urlcolor=coco3,
    citecolor=coco3,
}

\title{\vspace{-1cm}Endless Terminals: Scaling RL Environments \\for Terminal Agents}

\author{Kanishk Gandhi\thanks{Part of the work was done during a summer internship at Microsoft Research}\\
Stanford University
\And
Shivam Garg\\
Microsoft Research
\And
Noah D. Goodman\\
Stanford University
\And
Dimitris Papailiopoulos\\
Microsoft Research \\ UW-Madison
}

\begin{document}

\ifcolmsubmission
\linenumbers
\fi

\maketitle
\begin{abstract}\vspace{-3mm}

Environments are the bottleneck for self-improving agents. Current terminal benchmarks were built for evaluation, not training; reinforcement learning requires a scalable pipeline, not just a dataset. We introduce {\bf Endless Terminals}, a fully autonomous pipeline that procedurally generates terminal-use tasks without human annotation. The pipeline has four stages: generating diverse task descriptions, building and validating containerized environments, producing completion tests, and filtering for solvability. From this pipeline we obtain 3255 tasks spanning file operations, log management, data processing, scripting, and database operations. We train agents using vanilla PPO with binary episode level rewards and a minimal interaction loop: no retrieval, multi-agent coordination, or specialized tools. Despite this simplicity, models trained on Endless Terminals show substantial gains: on our held-out dev set, Llama-3.2-3B improves from 4.0\% to 18.2\%, Qwen2.5-7B from 10.7\% to 53.3\%, and Qwen3-8B-openthinker-sft from 42.6\% to 59.0\%. These improvements transfer to human-curated benchmarks:  models trained on Endless Terminals show substantial gains on held out human curated benchmarks: on TerminalBench 2.0, Llama-3.2-3B improves from 0.0\% to 2.2\%, Qwen2.5-7B from 2.2\% to 3.4\%, and Qwen3-8B-openthinker-sft from 1.1\% to 6.7\%, in each case outperforming alternative approaches including models with more complex agentic scaffolds. These results demonstrate that simple RL succeeds when environments scale. \footnote{Code available at \href{https://github.com/kanishkg/endless-terminals}{https://github.com/kanishkg/endless-terminals}}
\end{abstract}
\vspace{-6mm}
\begin{figure}[b]
    \centering
    \includegraphics[width=0.97\linewidth]{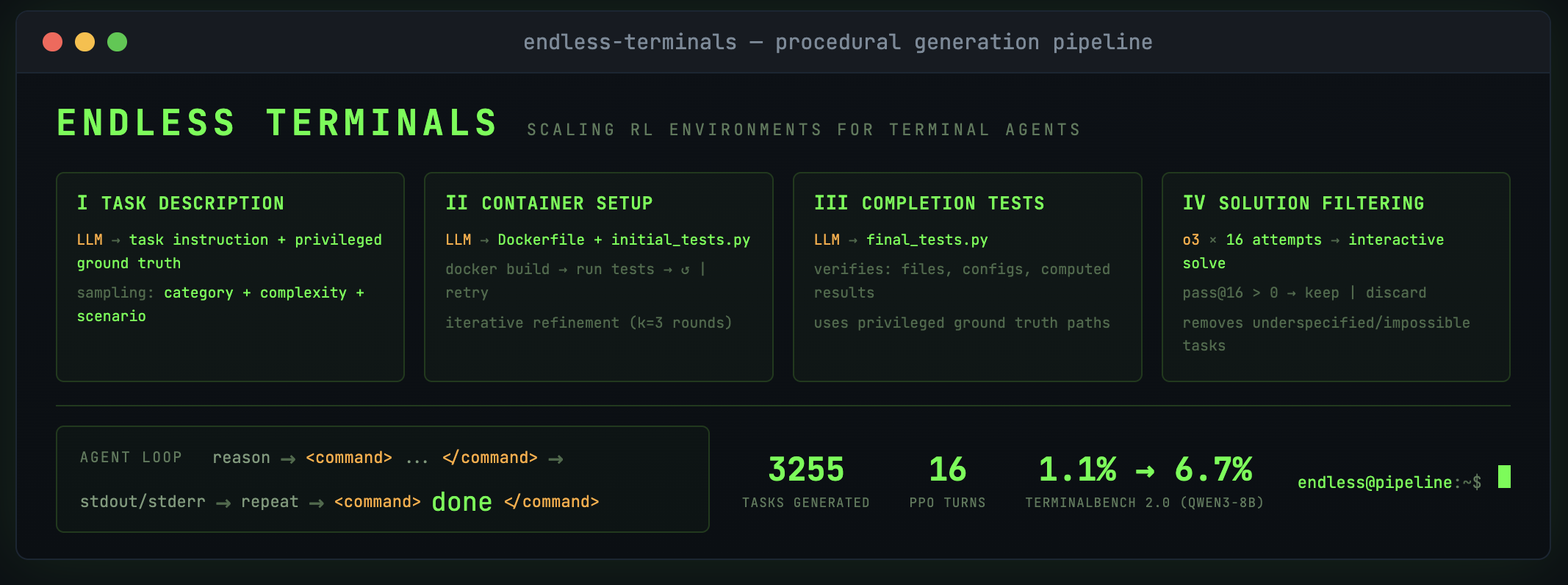}
    \vspace{-3mm}
    \caption{\textbf{Endless Terminals.} Tasks are procedurally generated through four phases: (I) task description generation, (II) container setup with iterative validation, (III) completion test generation, and
  (IV) solution-based filtering using o3. The pipeline yields 3255 verified tasks for training terminal agents with PPO.}
    \label{fig:placeholder}
\end{figure}
\vspace{-3mm}
\section{Introduction}

Reinforcement learning is hungry for environments. Successes of RL in improving language model reasoning, from mathematical problem solving to code generation have relied on access to a large set of diverse, and automatically verifiable tasks. Yet for training LLM agents to perform multi-turn computer tasks in a terminal, no such scalable environment exists. Real world terminal use requires agents to reason across multiple interactions, recover from errors, and  interactively execute sequences of commands that transform the stat of a system. Manual curation of such environments is expensive, and existing benchmarks offer at most hundreds of tasks, far too few to support robust RL training \citep{tbench_2025, primeintellect2025environments, openthoughts-agent}.

Prior approaches to building capable terminal agents have largely sidestepped this environment bottleneck. Some rely on fixed evaluation benchmarks repurposed for training, risking overfitting to narrow task distributions. Others distill behavior from stronger proprietary models through supervised finetuning \citep{guha2025openthoughtsdatarecipesreasoning}, inheriting the ceiling of the teacher and requiring expensive API access. A third line of work \citep{openthoughts-agent, lin2018nl2bash} uses human curated datasets of coding or shell tasks, but the annotation cost limits scale and diversity. What remains missing is a fully autonomous pipeline that can generate an endless stream of terminal tasks: complete with initial environments, task specifications, and verification tests, with minimal human supervision.

\begin{figure}[t]
    \centering
    \includegraphics[width=1.0\linewidth]{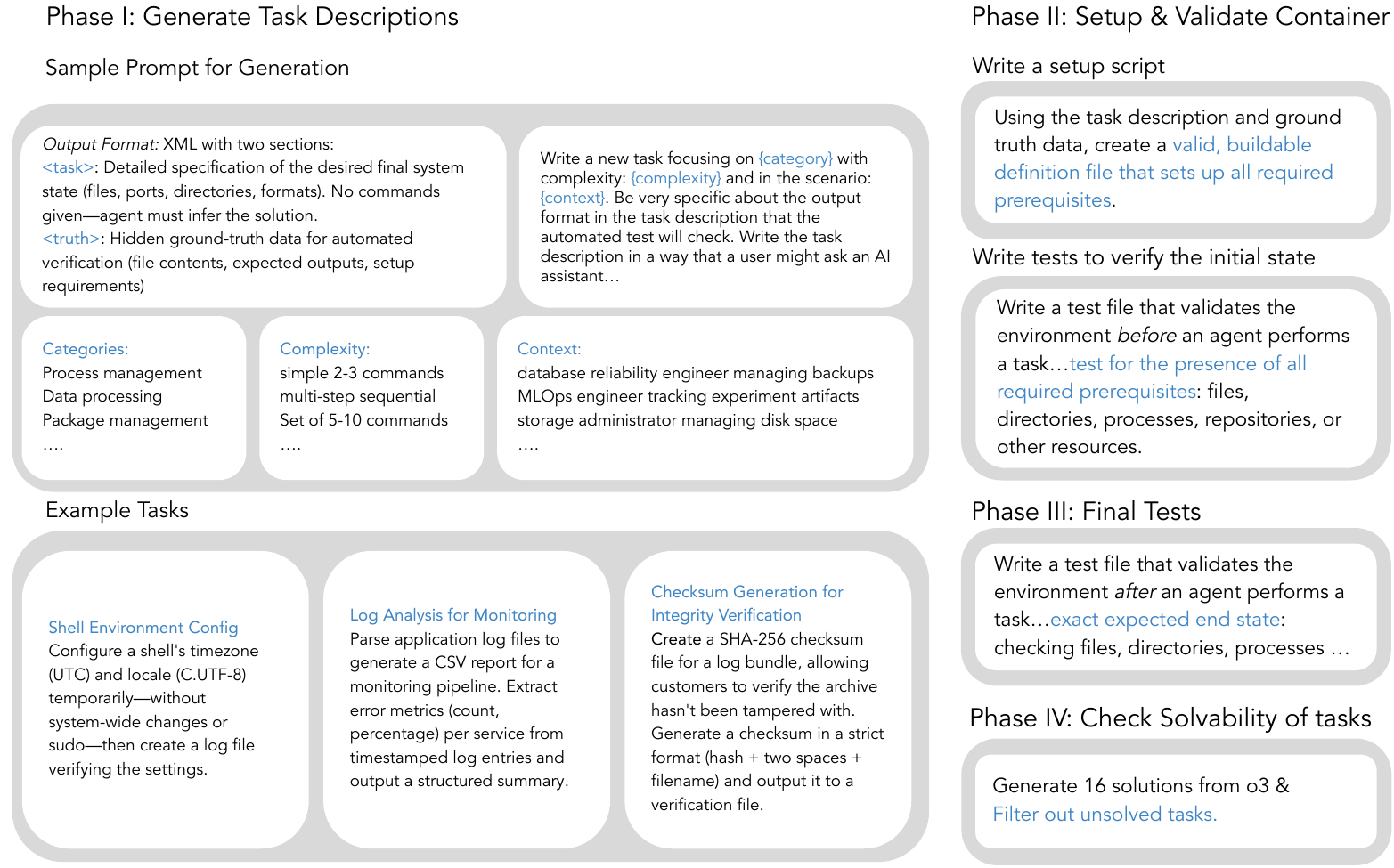}
    \caption{ \textbf{Overview of the Endless Terminals procedural generation pipeline.} Phase I generates diverse task descriptions by sampling across categories, complexity levels, and scenario contexts, producing both a task description and privileged ground truth data for verification. Phase II creates containerized environments and validates them with self written prerequisite tests. Phase III generates completion tests that verify the expected end state. Phase IV filters tasks by sampling 16 solutions from o3, retaining only tasks where at least one solution succeeds. Some example tasks include shell configuration, log analysis, and checksum generation.}
    \label{fig:pipeline}
\end{figure}

We introduce Endless Terminals, a procedural generation pipeline that synthesizes terminal-use tasks without human annotation or distillation. The pipeline operates in four stages (see \autoref{fig:pipeline}): 1) generating diverse task descriptions by sampling across categories, complexity levels, and scenario contexts; 2) building containerized environments and validating them with automatically generated prerequisite tests; 3) generating completion tests that verify the expected end state; and 4) filtering tasks by sampling solutions from a capable model to ensure solvability. 

We train models with a minimal agent architecture, a simple interaction loop where the model reasons, executes commands, and observes the output every turn, with no retrieval, tool use, or multi-agent scaffolding. Vanilla PPO \citep{schulman2017proximal} with 16 turns on Endless Terminals yields substantial gains: on our held-out dev set, Llama-3.2-3B \citep{grattafiori2024llama} improves from 4.0\% to 18.2\%, Qwen-2.5-7B \citep{qwen2024qwen2}  from 10.7\% to 53.3\%, and Qwen-3-8B-openthinker-sft \citep{yang2025qwen3,openthoughts-agent} from 42.6\% to 59.0\%. These gains transfer to difficult, human curated benchmarks. On TerminalBench 2.0 \citep{tbench_2025}, Llama-3.2-3B improves from 0.0\% to 2.2\%, Qwen-2.5-7B from 2.2\% to 3.4\%, and Qwen-3-8B-Open-Thoughts from 1.1\% to 6.7\%, in each case outperforming other finetuned versions of the corresponding models on TerminalBench.

These results demonstrate that an autonomous pipeline to scale environments can lead to improvements on human curated benchmarks and that simple RL setups succeed if we can scale them up.

\section{Related Work}

\paragraph{Agentic Scaffolds.} Prior work develops scaffolds for LLM agents with custom tools, retrieval, context management, and multi-agent coordination. SWE-agent \citep{yang2024sweagent} provides specialized commands for code navigation and editing. OpenHands \citep{wang2024openhands} combines an agent with tools for bash execution, file editing, and browser interaction. Terminus \citep{tbench_2025} takes a simpler approach, giving the agent only an interactive tmux session controlled via keystrokes. Our scaffold is simpler still: the agent reasons and emits commands, with the full history of prior thoughts, actions and shell outputs in context.

\paragraph{Supervised Finetuning and Distillation} One approach to building capable agents is to curate high quality datasets for supervised finetuning.  \citet{guha2025openthoughtsdatarecipesreasoning} (OpenThoughts) distill reasoning traces from frontier models, showing that data quality drives downstream improvement with RL. \citet{gandhi2025cognitive} identify cognitive behaviors: verification, backtracking, subgoal setting that enable self-improvement via RL, and show these can be instilled. \citet{olmo2025olmo}  demonstrate how midtraining on targeted data mixes to elicit capabilities like math and coding aids improvement with RL. Finally, OpenThinker-Agent \citep{openthoughts-agent} applies this recipe to terminal tasks, distilling traces from strong teachers. These approaches are complementary: SFT can provide a warm start for RL training.

\paragraph{Benchmarks and Interactive Environments.} Progress on coding agents has been driven by human curated benchmarks with execution based evaluation, including SWEBench \citep{jimenez2023swe} for GitHub issue resolution and TerminalBench 2.0 \citep{tbench_2025} for terminal tasks. These benchmarks require multiturn interaction where agents issue commands and observe outputs before deciding the next action, a format shared by  InterCode \citep{yang2023intercode} for interactive coding, and WebArena \citep{zhou2023webarena} for web navigation. Our procedurally generated tasks follow this multiturn format, with agents interacting with a persistent shell.

\paragraph{Synthetic Environment Generation} Some recent works construct environments for training agents on verifiable tasks. SWEGym \citep{pan2024training} provides 2438 Python tasks with executable tests, but relies on existing GitHub issues rather than procedural generation. In single turn domains,  \citet{poesia2024learning} propose models that posit and solve increasingly difficult problems training them through self-play. OpenThoughts Agent \citep{openthoughts-agent} is closest to our work, introducing terminal-use tasks for SFT and RL. However, their RL dataset involves human generated queries and commands from NL2Bash \citep{lin2018nl2bash}, and doesn't yield gains on benchmarks like Terminal Bench 2.0, or an autonomously generated endless terminals dev set. Endless Terminals generates tasks fully autonomously at arbitrary scale, and vanilla PPO yields substantial improvements that transfer to held out benchmarks.




\section{Procedural Generation of Tasks}

Our procedural generation pipeline (\autoref{fig:pipeline}) for EndlessTerminals consists of four stages: 1) generating task descriptions, 2) setting up an environment while validating it with self-written tests, 3) generating tests to verify completion of a task, and 4) generating several solutions from a strong model to ensure the validity of a task. Each stage builds on the previous, with automatic verification ensuring validity at every step.

\paragraph{Generating Task Descriptions} We prompt (\autoref{fig:pipeline}, Phase I) a language model to generate task descriptions paired with privileged ground truth information. To ensure diversity, each prompt samples randomly from three dimensions: task categories (like file management, text processing, log analysis, git operations, database queries, security scanning, among others), complexity levels (from single commands to multi-step sequences), and scenario contexts (like developer organizing files, DevOps engineer debugging logs, data analyst processing CSVs). The model outputs a task instruction, written as a request someone might pose to an AI assistant, along with a separate privileged information section containing exact file contents, paths, and expected states that automated tests will use for verification. The privileged information section is never revealed to the agent interacting with the environment.

\paragraph{Setting up and validating containers} Given the task description and ground truth, we generate two files (\autoref{fig:pipeline}, Phase II): an initial state test file that validates the container before the agent begins, and an Apptainer container definition or Dockerfile that establishes the required environment. The initial state tests verify prerequisites for the task: the presence of specific files, directories, running processes, or cloned repositories. For container generation, we employ an iterative refinement loop: the model generates a container definition, we build it and run the initial tests inside, and if tests fail, we feed the failure output back to the model for correction. This continues for up to $k=3$ rounds or until tests pass. Tasks that cannot produce a valid container are discarded.

\paragraph{Final Test Generation.} We generate a second test file that validates the system state after successful task completion. Given the task description, ground truth, and initial state tests, the model produces tests verifying expected outcomes, for example checking for the created files with correct contents, modified configurations, computed results (\autoref{fig:pipeline}, Phase III). These tests use paths, instructions and explicit data derived from the privileged ground truth. We verify that these tests do not pass in the initial state, ensuring they meaningfully assess task completion rather than trivially succeeding.

\paragraph{Solution Based Filtering.} To ensure tasks are solvable, we sample $n=16$ solution attempts from a capable model (o3 \citep{openai2025o3o4mini}) using our agent framework (\autoref{fig:pipeline}, Phase IV). Each attempt consists of an interactive session where the agent issues commands, observes outputs, and continues until declaring completion or exhausting its action budget. We retain tasks where at least one solution succeeds (pass@16 $> 0$) and discard the rest (see \autoref{fig:dist}, right). This filtering removes under specified or impossible tasks while confirming that retained tasks are achievable by strong models.

The complete pipeline processes tasks in parallel, failed tasks are discarded automatically. This approach gives us a controlled, automated and procedural way to generate tasks in a target domain while remaining automatically verifiable.

\section{Interacting with the Terminal}

\paragraph{Interaction Loop.}  At each turn, the model receives the conversation history, including its own previous reasoning and the shell outputs from prior commands, and produces either a command to execute or a signal that the task is complete. We use minimal XML tags to structure outputs: \texttt{<command>...</command>} wraps shell commands, and \texttt{<command>done</command>} indicates task completion. The model can include arbitrary reasoning before its command, which becomes part of the conversation history visible in subsequent turns. This means that the model can reference prior reasoning, correct mistakes, or build on partial progress.

\paragraph{Shell Environment.} For the shell, we support two types of containers: Docker and Apptainer. For Docker, we use the harbor framework \citep{Harbor_Framework_Team_Harbor_A_framework_2026}. For Apptainer, we maintain a persistent interactive shell session using a pseudo-terminal (PTY). The agent connects to an Apptainer container instance that remains alive across all turns of an episode, preserving filesystem state, environment variables, and running processes between commands. Each command executes in this persistent context. We capture both stdout and stderr, along with the exit code, and return a structured observation: whether the command succeeded or failed, followed by the output. This feedback is appended to the conversation as the next user message, and the loop continues.

\paragraph{Minimal Scaffolding.} The system prompt contains only basic instructions: output one command per turn, use non-interactive flags, verify solutions before declaring completion. So, tools like vim, htop etc. cannot be used by the agent.

\paragraph{Episode Termination.} An episode ends when the agent emits the done action, reaches a maximum number of turns (16 while training) or tokens (16k while training). We execute the held-out final tests inside the container to determine success.

So, we get a minimal loop between the model and the environment. The model reasons, acts, and observes, the conversation history accumulates, and the model reasons again with full context of what it has tried.

\section{Experiments}
\begin{figure}[tbp]
    \centering
    \includegraphics[width=1.0\linewidth]{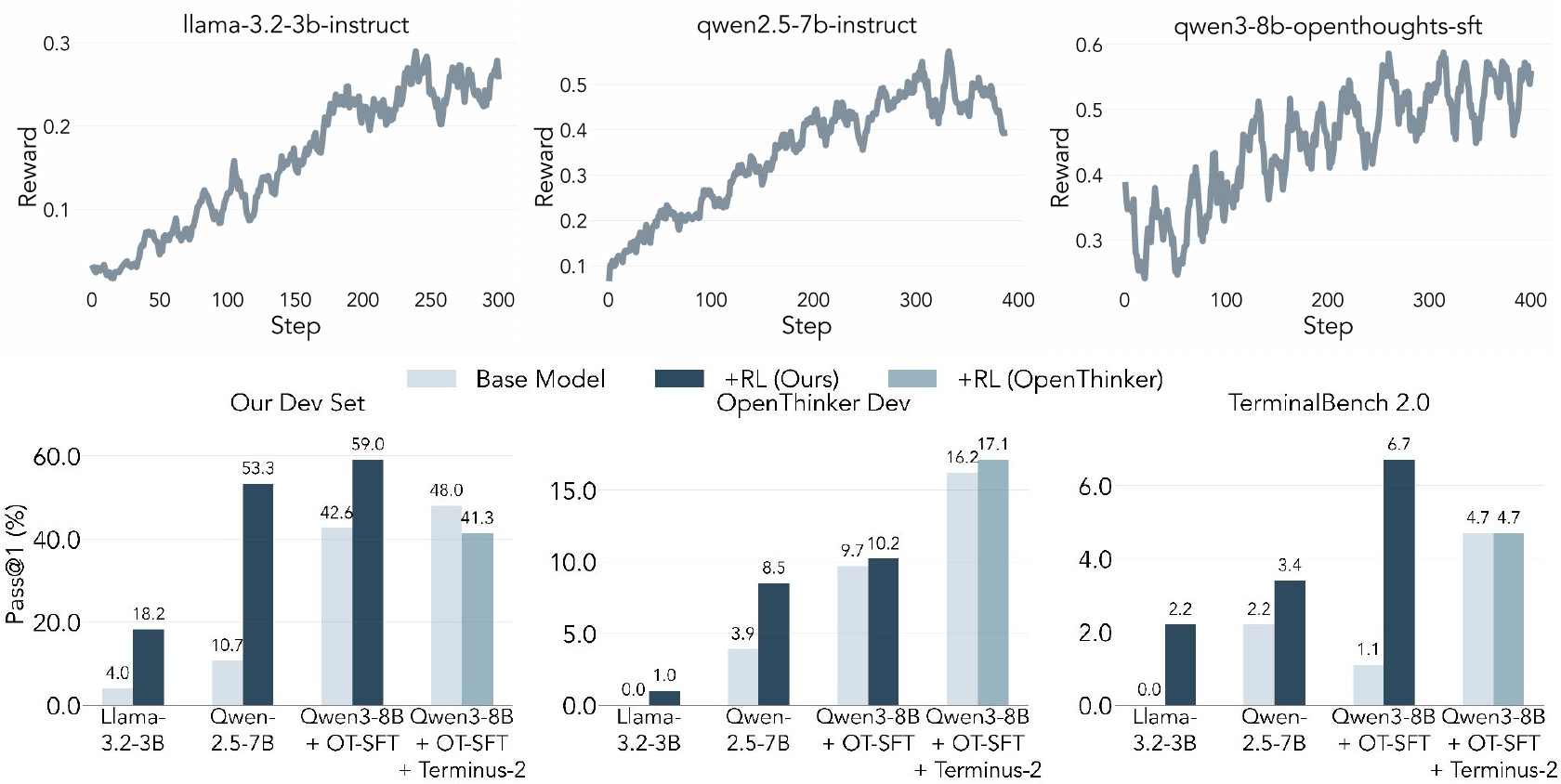}
    \caption{\textbf{Training and evaluation results.} Top row:  Reward curves during PPO training on Endless Terminals for (left) Llama-3.2-3B, (center) Qwen2.5-7B, and (right) Qwen3-8B-openthinker-sft, showing consistent improvement across all models. Bottom row: Pass rates on (left) our development set, (center)  OpenThinker development set , and (right) TerminalBench 2.0 . Models trained with our RL approach (+RL Ours) outperform base models and alternative finetuned variants across all evaluations. Here, RL (OpenThinker) denotes RL training in OpenThoughts-Agent \citep{openthoughts-agent}. Results on Terminal Bench 2.0 are averaged over 5 runs for our methods.
}
    \label{fig:results}
\end{figure}

\begin{wrapfigure}[21]{r}{0.5\textwidth}
    \centering
    \includegraphics[width=0.5\textwidth]{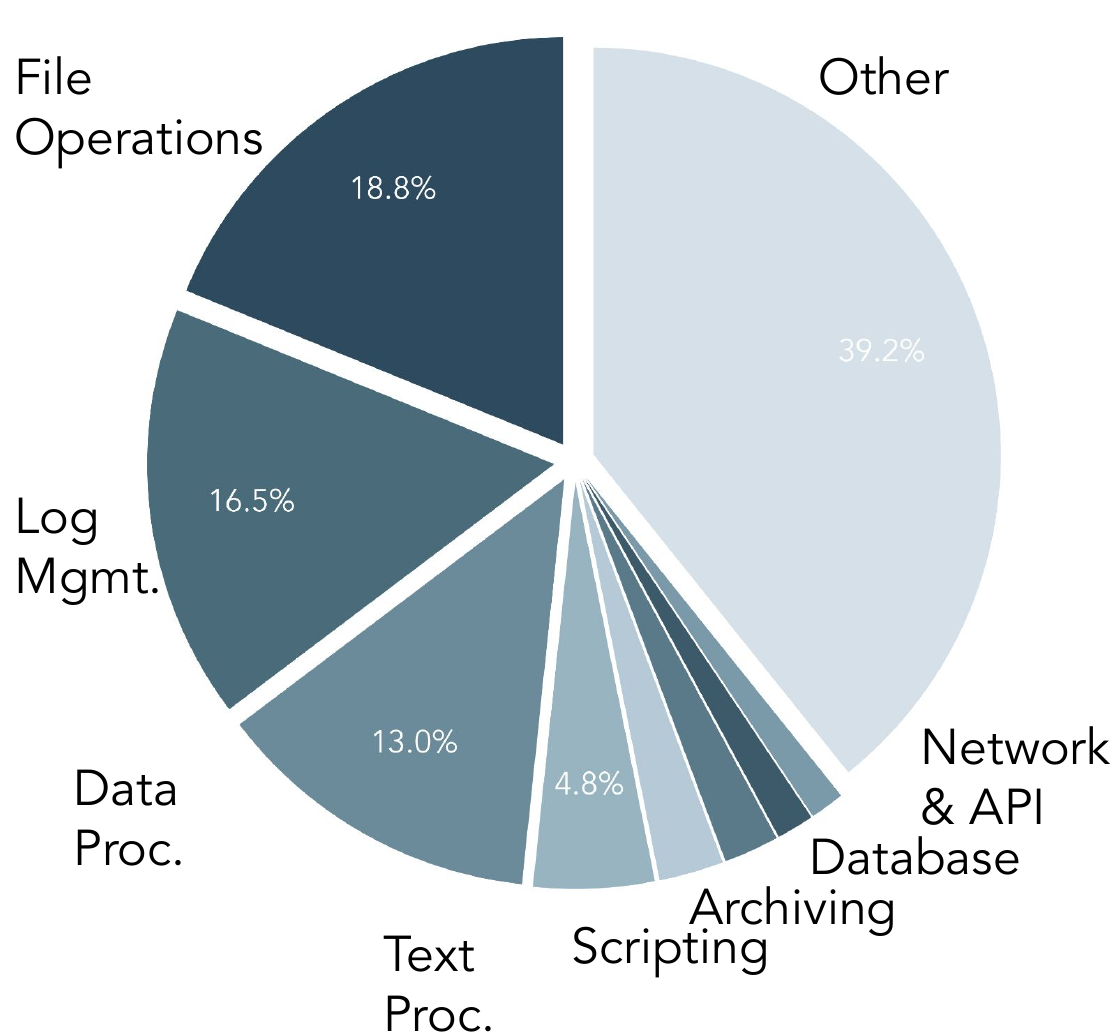}
    \vspace{-7mm}
    \caption{\textbf{Distribution of tasks in Endless Terminals.} Left: Task categories, with file operations and log management comprising the largest shares.}
    \label{fig:dist-pie}
\end{wrapfigure}

\textbf{Task Generation.} Our pipeline produces 3255 tasks in Apptainer format, of which approximately 2500 are also converted to Harbor \citep{Harbor_Framework_Team_Harbor_A_framework_2026} format. We use the Apptainer pipeline for all experiments. Solvability filtering discards roughly half of all generated candidates, those where o3 \citep{openai2025o3o4mini} fails all 16 attempts, removing underspecified or invalid tasks from the training distribution.

\begin{figure}[tbp]
    \centering
\includegraphics[width=0.95\linewidth]{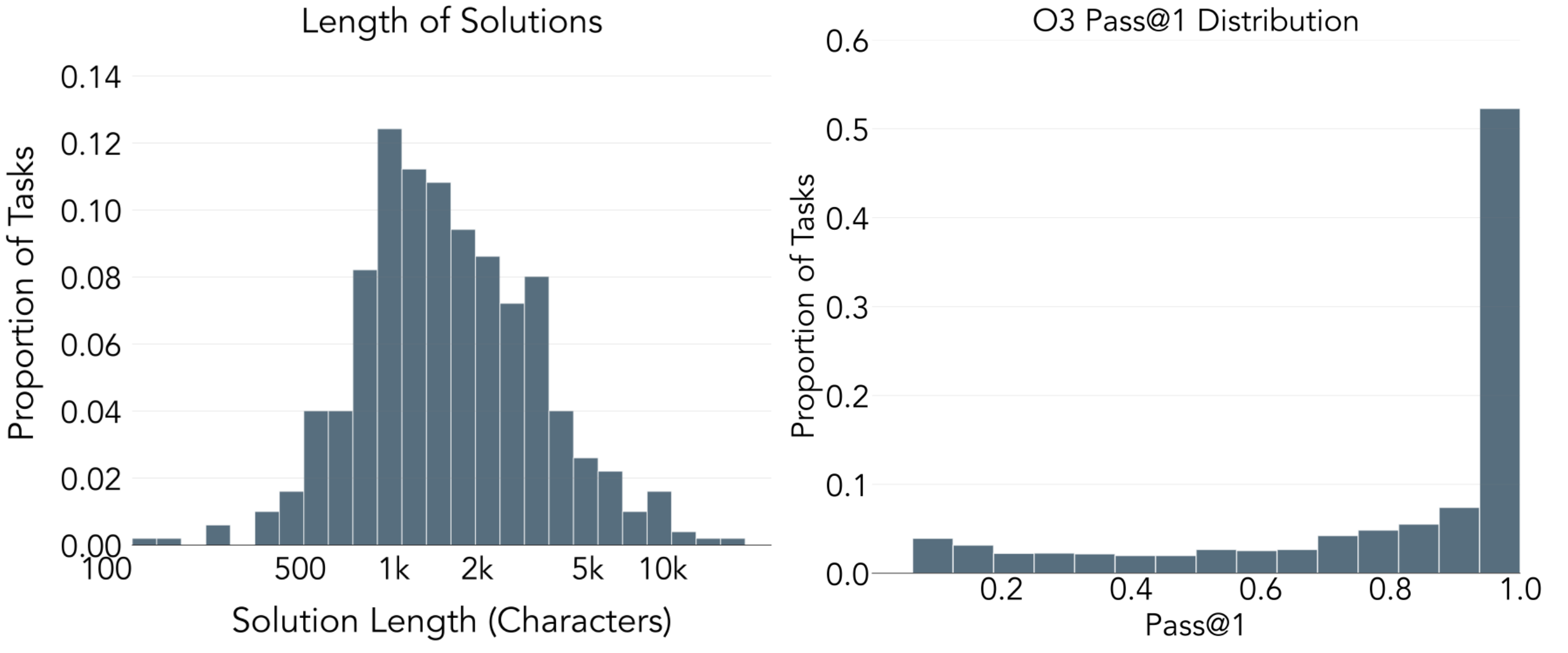}
    \caption{\textbf{Distribution of Generated Solutions.} (left) Distribution of solution lengths (in characters, log scale), showing most tasks require 1000-4000 characters of interaction with a long tail for complex tasks. (right) Distribution of pass rates across 16 o3 solution attempts, showing roughly half of tasks are solved by all attempts while the remainder span a range of difficulties.}
    \label{fig:dist}
\end{figure}

\paragraph{Endless Terminals produces a diverse distribution of solvable tasks.} \autoref{fig:dist-pie} shows the composition of our generated dataset. Task categories span file operations (the largest category), log management, data processing, text processing, scripting, archiving and compression and database operations. Solution lengths (\autoref{fig:dist}, left) vary considerably, with most tasks requiring between 1,000 and 4,000 characters of interaction, though the distribution has a long tail extending beyond 10,000 characters for more complex tasks.  The distribution of the success rate of solutions sampled from o3 (\autoref{fig:dist}, right) shows that roughly half the tasks are solved by all 16 attempts, with the remainder spanning a range of difficulties: though all are, by construction, within reach of current frontier models.

\paragraph{Training Setup.} We train our agent with Proximal Policy Optimization (PPO) \citep{schulman2017proximal}. Our implementation uses SkyRL \citep{griggs2025skrylv01}
For each training batch, we sample 16 rollouts per prompt for up to 16 turns, with a maximum of 2048 tokens generated per turn and a total context window of 16k tokens for the full conversation history. We use a temperature of 0.6 for both training and evaluation.

We treat each complete episode as a single reward signal: the agent receives reward 1 if the final tests pass and 0 otherwise, with no intermediate rewards. For the PPO objective, we use clipping bounds $\epsilon_{\text{low}} = 0.2$ and $\epsilon_{\text{high}} = 0.28$ \citep{yu2025dapo}, with sequence level loss averaging. We do not use a KL penalty since anecdotally we found that this hurt performance. To reduce the time for collecting rollouts, we add a 5 minute environment timeout. At inference time, we allow the model to interact for 64 turns, collapsing history when the context limit is reached by add the history of previous commands to the first user message. 

We train three models: Llama-3.2-3B-Instruct, Qwen2.5-7B-Instruct, and Qwen3-8B-openthinker-sft as our base models. Qwen3-8B-openthinker-sft is finetuned on 15000 traces from two sources: NL2Bash \citep{lin2018nl2bash}, synthetically generated tasks for shell command formatting, and InferredBugs \citep{inferfix2023}, a collection of C\# and Java bugs converted into interactive tasks. Traces are distilled from GLM-4.6 \citep{zeng2025glm}. Llama-3.2-3b and Qwen2.5-7b were trained on 4 A100s for about 2 days. Qwen3-8b-openthoughts-sft was trained on 8 B200s for about 8 hours.

\paragraph{Endless Terminals yields consistent improvement across models.} Training with PPO on procedurally generated tasks produces steady gains regardless of model size or starting capability. As shown in \autoref{fig:results} (top row), reward increases throughout training for all three base models—Llama-3.2-3B, Qwen2.5-7B, and Qwen3-8B-openthinker-sft. On our held-out development set, Llama-3.2-3B improves from 4.0\% to 18.2\%, Qwen2.5-7B from 10.7\% to 53.3\%, and Qwen3-8B-openthinker-sft from 42.6\% to 59.0\% This demonstrates that our procedural generation pipeline provides a reliable training signal.  On the OpenThinker development set, we observe similar trends: Llama-3.2-3B improves from 0.0\% to 1.0\%, Qwen2.5-7B from 3.9\% to 8.5\%, and Qwen3-8B-openthinker-sft from 9.7\% to 10.2\%. Gains on the OpenThinker development set are smaller, as this benchmark includes general software engineering tasks like issue resolution rather than purely terminal based tasks.

\paragraph{Simple RL succeeds when environments scale.} Our setup uses vanilla PPO with binary episode level rewards, no intermediate shaping, no KL penalty, and a minimal agent architecture without retrieval or multi-agent scaffolding. The gains come not from algorithmic sophistication but from scaling the environments, Endless Terminals provides the diverse, automatically verifiable tasks that RL requires.
\begin{wrapfigure}[18]{r}{0.45\textwidth}
    \centering
    \includegraphics[width=0.45\textwidth]{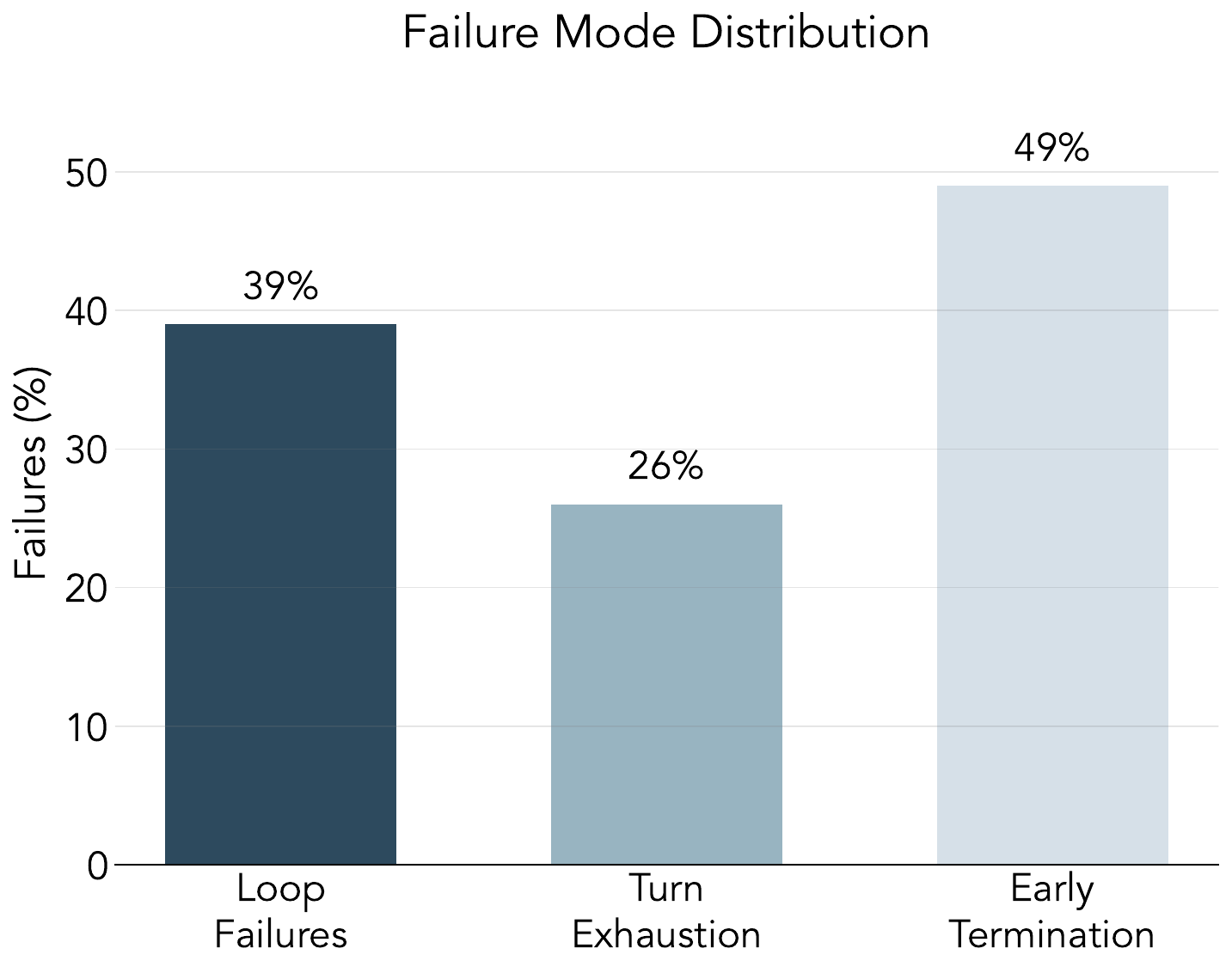}
    \vspace{-7mm}
    \caption{\textbf{Failure Analysis.} Failure modes are dominated by loop behaviors (39\%) and turn exhaustion (26\%), with remaining failures terminating early on specialized domains like cryptanalysis and bioinformatics (right).}
    \label{fig:failure}
\end{wrapfigure}

\paragraph{Gains transfer to held out human-curated benchmarks.} The improvements from training on Endless Terminals generalize beyond our procedurally generated distribution. On TerminalBench 2.0, a human curated benchmark the models never see during training, we observe substantial improvements: llama-3.2-3b improves from 0.0\% to 2.2\%, qwen-2.5-7b from 2.2\% to 3.4\%, and qwen-3-8b-open-thoughts-sft from 1.1\% to 6.7\%. In each case our RL trained models outperform other versions of the same base architectures (\autoref{fig:results}, bottom right), including models trained with alternative RL recipes \citep{openthoughts-agent} with agentic scaffolds like Terminus-2 \citep{tbench_2025}. Finally, our tasks were generated before the release of TerminalBench 2.0, ensuring no data leakage. 

\begin{figure}[btp]
    \centering
    \includegraphics[width=\linewidth]{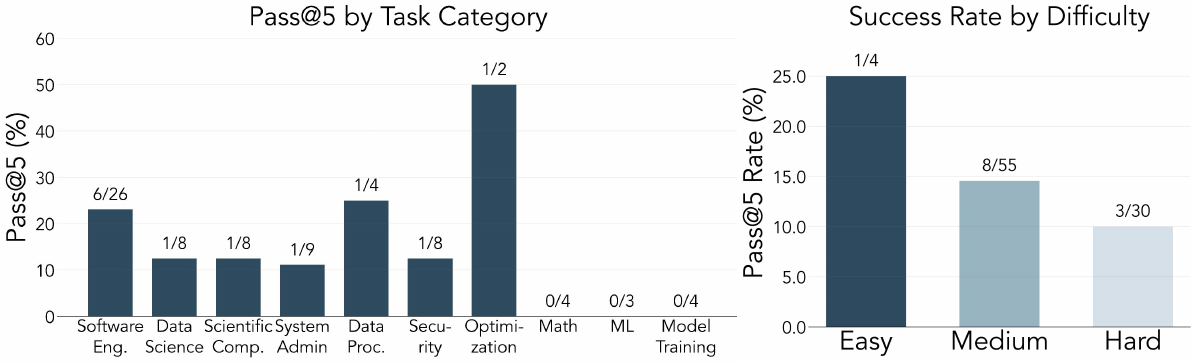}
    \vspace{-3mm}
    \caption{\textbf{Analysis of successes on TerminalBench 2.0.}  (right) vary substantially across categories, with software engineering tasks showing the best performance. and (left) Success rates (pass@5) decline with task difficulty }
    \label{fig:success}
\end{figure}

\paragraph{Failure Modes.} We analyze the tasks where our best model, qwen-3-8b-openthoguhts-sft achieved zero success on TerminalBench 2.0. For context, Claude Sonnet 4.5 with Terminus-2 achieves 42.8\% on this benchmark with a 200 turn limit and an agentic scaffold, compared to our 6.7\% with 64 turns and a no agentic scaffold. Pass@5 rates (at least one success in 5 attempts) decrease with task difficulty: 25\% (1/4) easy, 14.5\% (8/55) medium, and 10\% (3/30) hard tasks (\autoref{fig:success}).

We identify two primary failure modes (\autoref{fig:failure}): 1) loop failures \citep{pipis2025wait}, where the model becomes stuck repeating the same command sequence, accounting for 39\% of failures (30 tasks), and 2) turn exhaustion, where the model hits the turn limit (we train with 16 turns and evaluate with 64 turns with a sliding window), affecting 26\% of failures (20 tasks). These categories overlap, with 11 tasks exhibiting both behaviors. The remaining failures (49\%) terminate early with incorrect solutions, often in specialized domains like cryptanalysis, machine learning model extraction, and bioinformatics where the model lacks domain knowledge. 
 By category (pass@5), the model performs best on software-engineering (6/26), followed by data-science (1/8), optimization (1/2), scientific-computing (1/8), system-administration (1/9), data-processing (1/4), and security (1/8), while achieving zero success on mathematics (0/4), machine-learning (0/3), and model-training (0/4) tasks.
To understand loop failures, we measure \emph{command diversity}: the ratio of unique commands to total commands issued after the first error. Successful tasks exhibit significantly higher command diversity (0.49 on average) compared to failed tasks with loop behavior (0.18 on average), indicating that successful runs explore alternative approaches after errors while failed runs repeat commands.

\paragraph{Distillation and RL are complementary.} Our results show that starting from a stronger base amplifies gains from RL. Qwen3-8B-openthinker-sft, which was first finetuned on 15,000 distilled traces from NL2Bash and InferredBugs, achieves the highest final performance after RL training (6.7\% on TerminalBench 2.0). This suggests that SFT provides a warm start that RL can build upon more effectively \citep{gandhi2025cognitive, guha2025openthoughtsdatarecipesreasoning}.

\section{Discussion}

We introduced Endless Terminals, a procedural generation pipeline that synthesizes terminal-use tasks without human annotation or distillation from stronger models. The pipeline operates autonomously across four stages—task description generation, container setup with self, validation, completion test generation, and solution based filtering, producing 3,255 valid tasks spanning tasks like file operations, log management, data processing, text processing, etc. Training with PPO on these tasks yields consistent improvements across model scales, and these gains transfer to TerminalBench 2.0., a human curated benchmark. Our results demonstrate that simple RL setups can succeed when environments scale. Our failure mode analysis on TerminalBench 2.0 revealed two patterns: loop failures, where models repeat the same command sequences (39\% of failures), and turn exhaustion (26\%). 
Successful tasks exhibit significantly higher command diversity after the first error (0.49 vs 0.18 for looping failures), suggesting that exploring alternative approaches is crucial. Performance also varies by domain: software engineering tasks achieve 23\% success while mathematics, machine learning, and model training show zero success. This gap may reflect insufficient coverage of these domains in our procedural generation pipeline.

Our approach has several limitations worth noting. First, the procedurally generated tasks tend to resemble competitive programming problems more than the messy, underspecified requests that users actually pose to AI assistants. Real terminal use often involves ambiguous goals, implicit context, and might require clarifying questions. These are difficult to capture in automatically generated specifications without sacrificing verifiability. However, better user modeling \citep{shaikh2025creating,weston2025ai,gandhi2026learning} might enable building of such ``fuzzy'' environments. Conditioning the generation prompt to produce more naturalistic requests while maintaining sufficient specification for verification remains an open challenge.

Our filter for solvability introduces a capability ceiling. We filter using pass@16 from o3, retaining only tasks where at least one solution succeeds. This process that discards roughly half of all generated candidates. This filtering confirms that retained tasks are solvable and removes potentially underspecified or invalid tasks but also tasks that are beyond the capabilities of o3. This means that our pipeline cannot generate tasks beyond the frontier model's capability. As stronger models emerge, this ceiling will rise, but the dependence on a frontier validator limits our ability to train agents on truly novel problems. Self-play approaches, where models iteratively generate tasks just beyond their current capability and learn to solve them, could adaptively scale difficulty without relying on a fixed frontier validator \citep{poesia2024learning, zhao2025absolute}.

Incorporating humans in the loop, either to validate generated tasks or to provide naturalistic task descriptions, could improve both task quality and diversity beyond what purely synthetic generation achieves, albeit at the increasing the cost of generating tasks, making the pipeline less scalable. Testing RL with richer agentic scaffolds (retrieval, multi-agent coordination, tool use) may yield further gains. Partial rewards based on the number of test cases passed, rather than binary episode-level rewards, could provide denser training signal and accelerate learning \citep{sun2025rl}.  Finally, learning world models of terminal dynamics \citep{copet2025cwm} or distilling environment dynamics into reasoning-based experience models \citep{chen2025scaling} could enable more sample-efficient training by allowing agents to plan and simulate outcomes through imagined rollouts before executing commands.

These directions, task verification, richer scaffolds, denser reward signals, and learned world models, highlight that building capable terminal agents remains a multifaceted challenge. Endless Terminals represents one cog in this greater effort, demonstrating that scalable environment generation can unlock substantial gains even with simple RL setups. We hope this work encourages the community to invest in automated task generation pipelines alongside algorithmic advances, richer scaffolds, and improved training objectives.





\section*{Acknowledgments}
The authors would like to thank Vaish Shrivastava, Sahaj Agarwal, Vasilis Kontonis, Ahmed Awadallah, Corby Rossett and Shital Shah for discussions and their support. This work was started during KG's internship at Microsoft Research. KG was supported by an HAI-SAP Grant, the Affective Science Fellowship and an NSF Expeditions grant during his time at Stanford.

\bibliography{colm2025_conference}
\bibliographystyle{colm2025_conference}

\appendix

\end{document}